\documentclass[conference]{IEEEtran}
\usepackage{times}
\usepackage[numbers]{natbib}
\usepackage{multicol}
\usepackage{amsmath,amssymb}
\usepackage{graphicx}
\usepackage{booktabs}
\usepackage{xcolor}
\usepackage{caption}
\usepackage{dsfont}
\usepackage{multirow}  
\usepackage{algorithm}
\usepackage{nicefrac}
\usepackage{xfrac}
\usepackage{xspace}
\usepackage{subcaption}
\usepackage{amssymb}
\usepackage{amsfonts}
\usepackage{bbding}
\usepackage{tabularx} 
\usepackage{epigraph} 

\definecolor{cvprblue}{rgb}{0.21,0.49,0.74}

\usepackage[
  breaklinks=true,
  colorlinks=true,
  linkcolor=red,
  citecolor=cvprblue,
  filecolor=magenta,
  bookmarks=true,
]{hyperref}

\newcommand{\num}[1]{#1} 

\begin{document}

\title{UR-VC: Unsupervised Robotic Value Correction for Time-Derived Progress Proxies}

\author{

\authorblockN{
Lirui Zhao,
Modi Shi,
Li Chen,
Qi Liu,
Ping Luo,
Hongyang Li
}

\smallskip

\authorblockA{
The University of Hong Kong\\
\smallskip
\url{https://liruizhao.com/projects/UR-VC}
}
}

\maketitle

\begin{abstract}
Modern robot learning systems increasingly rely on dense progress or value signals to evaluate intermediate states, guide policy learning, and detect task completion, making the quality of these signals critical. Since such dense labels are rarely available at scale, normalized time within a demonstration is often used as a scalable substitute: later frames are treated as higher progress. However, this time-derived label is only a noisy proxy for physical task progress. In contact-rich manipulation, a robot may make progress and then lose it through slips, failed grasps, or partial undoing, while the time-derived label continues to increase monotonically.
We introduce Unsupervised Robotic Value Correction (UR-VC), an offline, training-free method for correcting time-derived progress labels. UR-VC exploits a simple regularity in demonstration data: similar states often recur across different episodes, but at different timestamps. Instead of trusting the timestamp from a single trajectory, UR-VC retrieves similar states from other episodes and aggregates their time-derived labels to obtain a corrected progress estimate. UR-VC requires no manual progress labels, reward annotations, or additional value model.
We evaluate UR-VC on real bimanual cloth flatten-and-fold data, a long-horizon deformable-object manipulation task with visible intermediate progress. The corrected labels capture local regressions and non-uniform progress that normalized time cannot represent, while preserving the overall task trend. We further use the corrected signal to construct advantage labels for VLA training, following recent advantage-conditioned policy learning. UR-VC shows a positive trend in real-robot task success under matched data, model, and training settings.

\end{abstract}

\IEEEpeerreviewmaketitle

\begin{figure*}[t]
\centering
\includegraphics[width=0.85\textwidth]{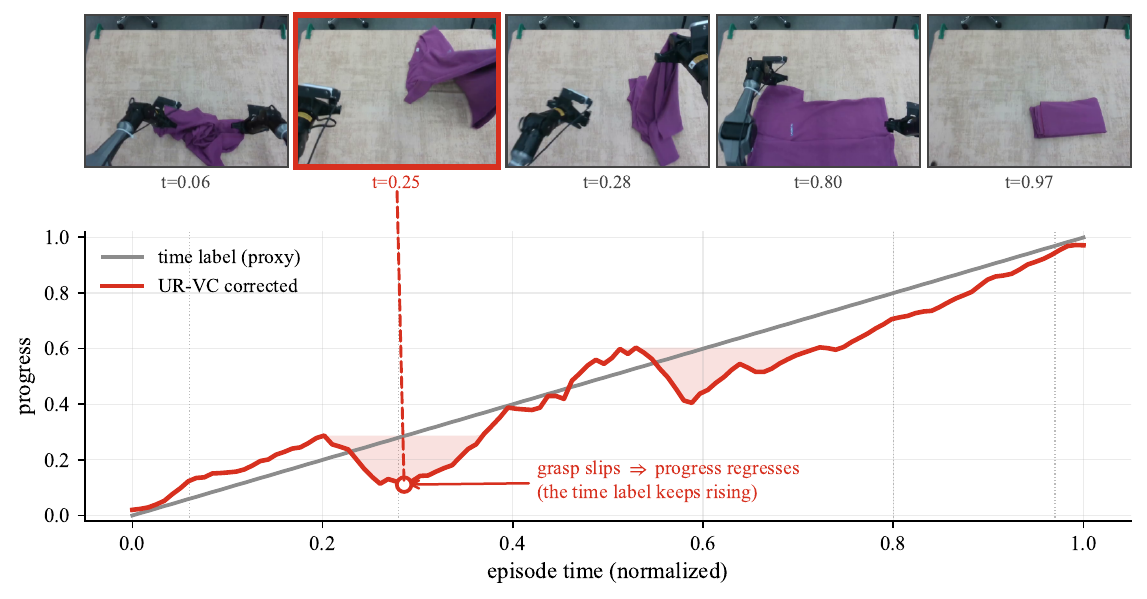}
\vspace{-1em}
\caption{\textbf{Time is not progress.}
In a real cloth flatten-and-fold episode, the garment is first smoothed, but a slipping grasp reintroduces wrinkles and causes physical progress to regress. The time-derived label (gray) remains monotone by construction, whereas UR-VC estimates corrected progress by retrieving semantically similar frames from other episodes and aggregating their time-derived labels, producing a correction (red) that decreases when the visual state worsens. Across the primary evaluation set, \num{13.4}\% of frames receive a negative horizon advantage under UR-VC, capturing regressions the time proxy misses.
}
\vspace{-1em}
\label{fig:teaser}
\end{figure*}

\section{Introduction}

Modern robot learning systems increasingly use intermediate task signals in addition to action supervision.
In long-horizon manipulation, such signals estimate how far the current state has advanced toward task completion and whether recent actions have moved the task forward.
They commonly take the form of task progress, value, or advantage, and are used for completion detection, value-function learning, and advantage-conditioned policy learning~\cite{intelligence2025pistar06,geminirobotics2025gr15,ma2025gvl,peng2019awr,chen2021dt,emmons2022rvs}. 
As vision–language–action (VLA) models are trained on larger and more diverse robot datasets~\cite{brohan2023rt2,kim2024openvla,black2024pi0,intelligence2025pi05}, the quality of these dense signals becomes increasingly critical.

The difficulty is that dense progress or value labels are rarely available at scale. A common scalable substitute is normalized time within a successful demonstration~\cite{intelligence2025pistar06,geminirobotics2025gr15}, which treats later frames as higher progress since they usually correspond to states closer to task completion. However, the resulting time-derived label is only a noisy proxy for physical task progress. It assumes that progress increases monotonically and uniformly with time, while real manipulation, especially contact-rich and deformable manipulation~\cite{lin2020softgym,deng2024clasp}, can regress, stall, or advance at different rates. For example, in the real cloth flatten-and-fold episode (Fig.~\ref{fig:teaser}), the robot first smooths the garment, but a later grasp slip reintroduces wrinkles and moves the task state backward, so physical progress regresses even as the time-derived label continues to increase linearly.

This mismatch matters for downstream learning in two ways.
Within an episode, this directly affects advantage-style supervision: under a fixed-horizon progress difference, a linear time proxy assigns the same increase to every frame, so it cannot distinguish actions that improve the physical state from actions that stall or undo progress.
Across episodes, the problem is subtle but more persistent: similar physical states can receive different labels simply because demonstrations proceed at different speeds or take different recovery motions.
A model trained on these noisy labels may then explain label variation using incidental cues such as appearance, texture, or execution speed, rather than physical task progress.

A common response is to learn this signal from data, for example by training progress estimators or value models using temporal contrast, language--image objectives, value pretraining, optimal transport, or generative modeling~\cite{sermanet2018tcn,ma2023liv,ma2023vip,bhateja2023vptr,fu2024temporalot,huang2024diffusionreward}, or by eliciting value estimates from pretrained models~\cite{ma2025gvl,topreward2026,lv2026viva}.
However, when the supervision target is a systematically biased time-derived proxy, a learned estimator can inherit the same bias: it may smooth over local regressions and encode differences in execution speed or recovery behavior as differences in progress.
This motivates a complementary question: \emph{can the proxy labels themselves be corrected before they serve as supervision for estimators or policies?}

We introduce \textbf{Unsupervised Robotic Value Correction (UR-VC)}, an offline, training-free method for correcting time-derived progress labels in demonstration datasets.
UR-VC is based on a simple data-driven premise: time-derived labels are noisy mainly because each demonstration has its own temporal distortion. 
Cross-episode state matches provide anchors for reducing this distortion: when similar physical states are found in independently collected episodes, the state is approximately held fixed while the timestamp varies across trajectories.
Aggregating the time-derived labels of these matched states therefore estimates where that state typically lies in the task, rather than where it happened to occur in one particular episode.
UR-VC operationalizes this idea by retrieving semantically similar frames from other episodes using SigLIP-2 visual embeddings~\cite{tschannen2025siglip2}
and aggregating their time-derived labels into a corrected progress estimate, with at most one matched label contributed by each episode to avoid over-weighting temporally adjacent frames.

We study UR-VC on real bimanual cloth flatten-and-fold data. Cloth manipulation is a natural testbed for this problem because progress is visually meaningful, local regressions occur frequently, and similar intermediate states recur across garments and episodes~\cite{zhao2023aloha,deng2024clasp}.
Since true physical progress is not directly observed in demonstrations, we first evaluate the conditions that make correction useful: cross-episode coverage, task-state consistency, stability under aggregation, and recovery of non-monotone progress. 
We find that these conditions hold in our data: high-quality matches are abundant even at moderate dataset sizes (Fig.~\ref{fig:matching}), retrieved frames align with folding state across different garments (Fig.~\ref{fig:montage}), the corrected estimate becomes smoother as the evaluation set grows (Fig.~\ref{fig:smooth}), and the labels recover local regressions while preserving the overall task trend (Fig.~\ref{fig:teaser}).
As a downstream use case, we convert the corrected signal into advantage labels for VLA training, following recent advantage-conditioned policy learning~\cite{intelligence2025pistar06,peng2019awr}, and observe a positive trend in real-robot task success under matched training settings (Tab.~\ref{tab:real}).

Our contributions are summarized as follows:
\begin{itemize}
\item \textbf{Proxy-label perspective.} 
We formulate normalized time as a noisy proxy for latent physical task progress, and highlight its limitations both within episodes and across episodes in non-monotone manipulation.
\item \textbf{Unsupervised correction.} 
We propose UR-VC, an offline and training-free method that corrects time-derived progress labels by aggregating semantically similar states across episodes, requiring no manual progress labels, reward annotations, or an additional value model.
\item \textbf{Real-robot evidence.}
On real bimanual cloth flatten-and-fold data, we validate the conditions needed for correction and demonstrate a downstream use case in advantage-conditioned VLA training.
\end{itemize}

\section{Related Work}
\label{sec:related}

\textbf{Progress and value supervision in robot learning.}
Vision-language-action models such as RT-2~\cite{brohan2023rt2}, OpenVLA~\cite{kim2024openvla}, and the \(\pi_0\)/\(\pi_{0.5}\) family~\cite{black2024pi0,intelligence2025pi05} make large-scale robot policy learning practical, and recent systems consume explicit progress or value signals: \(\pi^{*}_{0.6}\)~\cite{intelligence2025pistar06} learns a value function for advantage conditioning, and Gemini Robotics~1.5~\cite{geminirobotics2025gr15} uses stage progress for planning and completion judgment. A body of work learns visual progress, reward, or value estimators from contrastive video objectives, language-image representations, value pretraining, optimal transport, or generative modeling~\cite{sermanet2018tcn,sontakke2023roboclip,ma2023liv,ma2023vip,bhateja2023vptr,fu2024temporalot,huang2024diffusionreward}, or elicits values from large pretrained models~\cite{ma2025gvl,topreward2026,lv2026viva}. UR-VC is complementary to both lines: it corrects the proxy scores \emph{before} any estimator is trained, targeting the supervision itself.

\textbf{Cross-episode redundancy and label correction.}
UR-VC is also related to learning with noisy labels~\cite{natarajan2013learning,frenay2014classification} and graph/nearest-neighbor label propagation~\cite{zhu2003semi,iscen2019label}. The difference is the source of redundancy. Rather than propagating sparse human labels over an image graph, UR-VC exploits repeated robot states across independently collected episodes, where each timestamp is a noisy measurement of the same latent physical progress.

\textbf{Deformable manipulation and semantic encoders.}
Cloth manipulation has been studied in simulation and real-robot settings, including SoftGym~\cite{lin2020softgym}, ALOHA/ACT~\cite{zhao2023aloha}, Mobile ALOHA~\cite{fu2024mobilealoha}, and CLASP~\cite{deng2024clasp}.
UR-VC studies real bimanual cloth flatten-and-fold data since this setting exposes both within-episode regressions and across-episode timing variation in time-derived progress labels. We use SigLIP-2~\cite{tschannen2025siglip2}, an image--text pretrained semantic vision encoder, to retrieve cross-episode state matches.

\section{Method}
\label{sec:method}

We first formalize normalized time as a noisy proxy for latent task progress and explain why its fixed-horizon difference provides a degenerate advantage signal. We then introduce UR-VC, which corrects this proxy by aggregating time-derived labels from semantically matched states across episodes. Finally, we describe how the corrected progress estimate is converted into advantage labels for policy training.

\subsection{Time-derived progress as a noisy proxy}
\label{sec:proxy}

Let episode \(e\) contain frames \(o^{(e)}_1,\dots,o^{(e)}_{T_e}\). A common time-derived proxy defines 
\begin{equation}
g^{(e)}_t=t/T_e,
\end{equation}
treating elapsed time as progress supervision. 
We assume there is also an unobserved task progress \(p(o)\in[0,1]\), determined by the physical state relevant to the task.
This progress \(p\) is a conceptual target, not a quantity available to the algorithm. 

The mismatch between \(g\) and \(p\) appears in two ways.
First, within an episode, physical progress can be non-monotone while normalized time is monotone by construction. For instance, if a cloth fold slips or a grasp fails, \(p(o^{(e)}_t)\) may decrease even though \(g^{(e)}_t\) continues to increase. As a result, raw time labels cannot distinguish actions that improve the task state from actions that stall or locally undo progress.
Second, across episodes, the same physical state can appear at different normalized times. Operators may move at different speeds, pause for different durations, or take different recovery paths after local failures. Thus, \(g^{(e)}_t\) contains not only information about task progress, but also episode-specific timing distortion. A frame that is physically close to completion may receive a lower time-derived label in a slow episode, while a less advanced state may receive a higher label in a fast episode.

These two issues directly affect downstream learning. For advantage conditioning with a fixed action horizon \(H\), the finite difference of the raw time proxy is
\begin{equation}
  g_{t+H} - g_t = \frac{H}{T_e},
\end{equation}
constant within episode \(e\). 
Therefore, within the same episode, naively differencing the raw label gives every frame the same ranking signal, regardless of whether the action improves or worsens the physical state.

Across episodes the difference varies only through the episode length \(T_e\), which primarily reflects execution speed rather than state improvement. More generally, any method that consumes raw time labels as progress or value supervision inherits these within-episode monotonicity errors and across-episode timing distortions.

\begin{figure}[t]
\centering
\includegraphics[width=\linewidth]{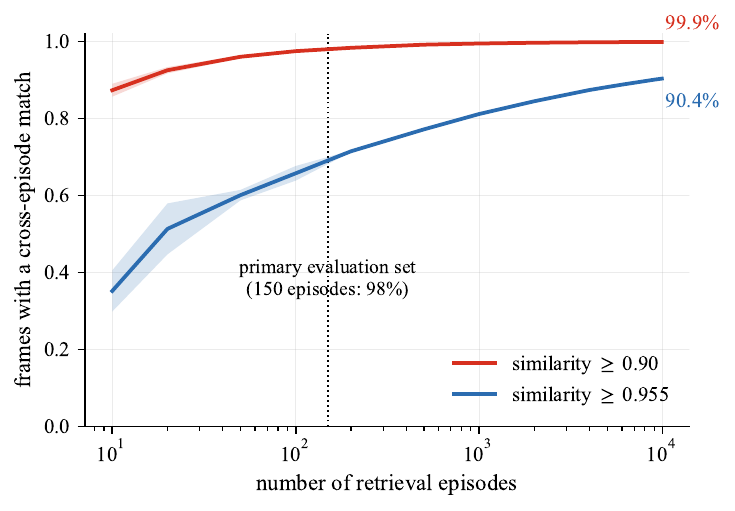}
\vspace{-1em}
\caption{\textbf{Cross-episode matches are abundant.}
Fraction of task-region frames with at least one semantically similar neighbor in a \emph{different} episode as more retrieval episodes are added. On real flatten-fold demonstrations, the \num{150}-episode primary evaluation set already covers \num{98}\% of frames at similarity \(\ge 0.90\), indicating that independently collected episodes provide sufficient anchors for correcting time-derived progress labels. Coverage further improves at the \(10^4\)-episode scale, reaching \num{99.9}\% at \(\ge 0.90\) and \num{90.4}\% at \(\ge 0.955\).}
\vspace{-1em}
\label{fig:matching}
\end{figure}

\begin{figure*}[t]
\centering
\includegraphics[width=0.92\textwidth]{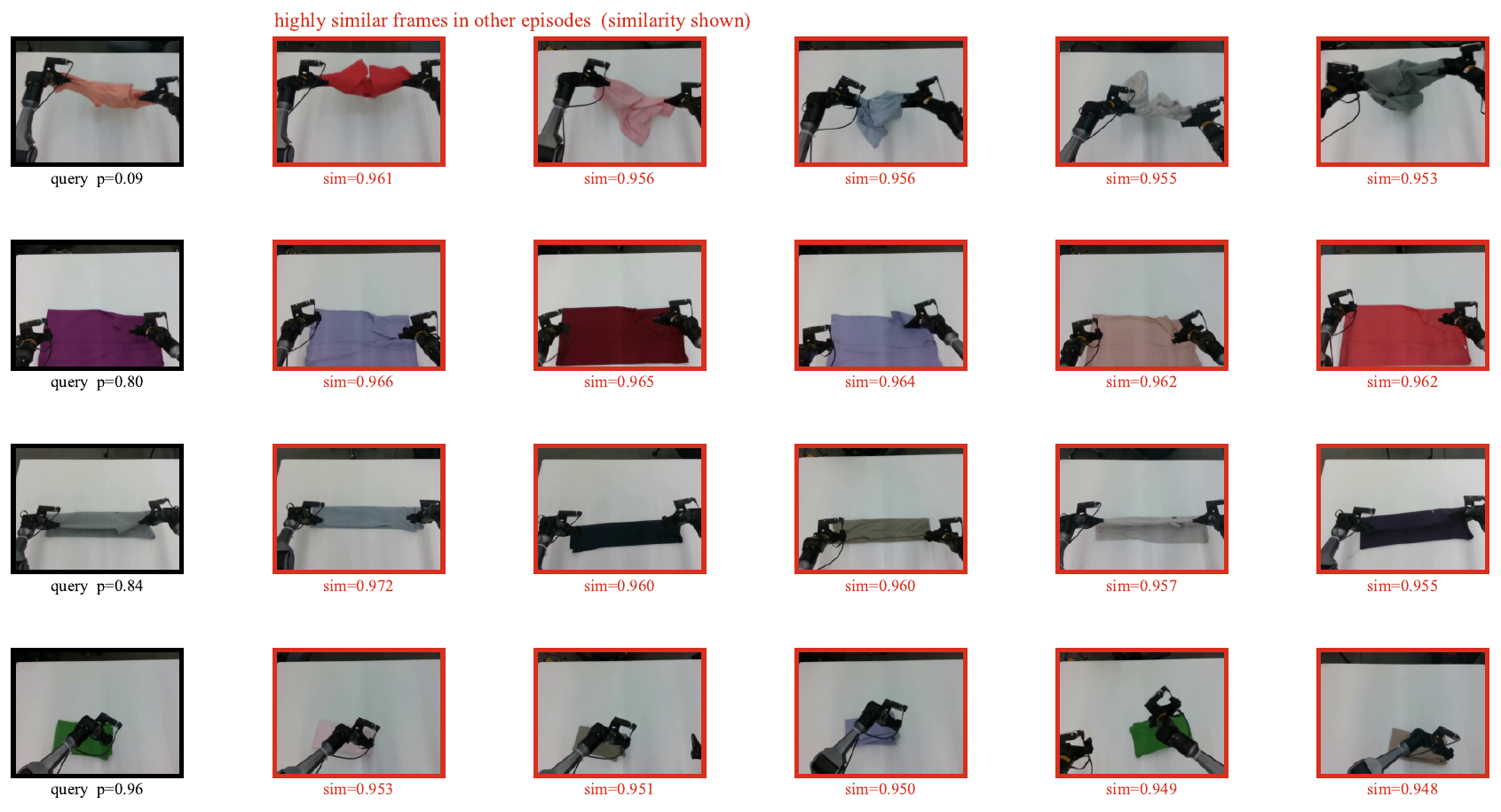}
\caption{\textbf{Cross-episode retrieval captures folding state across garment appearances.} 
For query frames at successive fold stages (left, black borders), nearest neighbors retrieved from \emph{other} episodes (red borders, similarity shown) exhibit the same folding state across different garment colors.
}
\vspace{-1em}
\label{fig:montage}
\end{figure*}

\subsection{UR-VC: Episode-Balanced Cross-Episode Correction}
\label{sec:correction}

UR-VC corrects a time-derived proxy by exploiting cross-episode recurrence: similar physical states often appear in independently collected demonstrations, but at different normalized times. For an idealized recurring state, episode \(e\)'s time-derived score can be viewed as
\begin{equation}
  g^{(e)} = p + \varepsilon_e,
\end{equation}
where \(p\) is the latent progress and \(\varepsilon_e\) is an episode-specific timing error. 
Averaging proxy scores across independent episodes can reduce this error; averaging many nearby frames from the same episode would not, since their errors are correlated.

Concretely, let \(f_i\) be the \(L_2\)-normalized SigLIP-2 embedding of query frame \(i\), and let \(f_i^\top f_j\) denote cosine similarity. For every other episode \(e\), we form an in-band candidate set
\begin{equation}
\mathcal{C}_{i,e}=\{\,j : \mathrm{ep}(j)=e,\ |g_j-g_i|\le\tau\,\}.
\end{equation}
We then keep only the best representative from that episode,
\begin{equation}
  j^*_{i,e}=\arg\max_{j\in\mathcal{C}_{i,e}} f_i^\top f_j,
\end{equation}
and discard it if its similarity is below \(\rho\). Let \(\mathcal{M}_i\) be the remaining episode representatives, optionally capped to the top \(m\) episodes by similarity. The corrected estimate is
\begin{equation}
  \hat g_i = \frac{1}{|\mathcal{M}_i|}\sum_{e\in\mathcal{M}_i} g_{j^*_{i,e}}.
  \label{eq:vc}
\end{equation}
If no representative survives, we leave \(g_i\) unchanged.

The episode-balanced form matches the statistical assumption above: each episode contributes at most one proxy score to the average, so the estimate aggregates evidence across demonstrations rather than across temporally adjacent frames from a single trajectory. 
The similarity threshold \(\rho\) acts as a conservative match filter, preventing episodes without a sufficiently similar state from contributing to the estimate. The temporal band \(\tau\) imposes a locality constraint on label correction: all averaged labels satisfy \(|g_j-g_i|\le\tau\), and therefore the correction magnitude is bounded as \(|\hat g_i-g_i|\le\tau\).

UR-VC is an offline label-correction procedure. After computing visual embeddings, it only requires similarity search, per-episode representative selection, and scalar averaging. In implementation, the per-episode argmax is computed with a single masked scatter-max over the offline corpus, so the correction reduces to a small number of matrix operations. UR-VC doesn't require manual progress labels, reward annotations, online rollouts, or training an additional value model.

\subsection{Advantage Labels from Corrected Progress}
\label{sec:application}

After correcting the time-derived proxy, we use the resulting estimate \(\hat g\) to construct advantage labels for policy training following \(\pi^{*}_{0.6}\). This step is only one downstream use of UR-VC: the correction is performed at the progress-label level, before any policy or value model is trained.

For an action chunk starting at frame \(i\) with horizon \(H\), we define a scalar corrected advantage proxy as
\begin{equation}
  r_i = \hat g_{i+H}-\hat g_i.
\end{equation}
Near the end of an episode, we compute the difference to the final frame and rescale it to the horizon rate:
\begin{equation}
r_i =
\frac{H}{T_e-i}
\bigl(\hat g_{T_e}-\hat g_i\bigr).
\label{eq:end_rescale}
\end{equation}
Here we follow the indexing convention \(o^{(e)}_1,\dots,o^{(e)}_{T_e}\). If the implementation uses zero-indexed frames, the denominator should be adjusted accordingly.

We rank frames by \(r_i\) and mark the top \(\num{20}\%\) as positive-advantage examples. The corresponding training frames receive a plain-text prompt suffix, e.g., ``\texttt{…, advantage: positive}’’. At deployment, the policy is queried with the same positive-advantage suffix. The policy architecture, training data, and optimization schedule are otherwise unchanged.

Thus, UR-VC introduces no new policy objective and no separate value model. It only replaces the raw time-derived supervision with a corrected signal, allowing existing progress- or advantage-conditioned pipelines to use cross-episode information without additional annotation or critic training.

\section{Experiments}
\label{sec:experiments}

We evaluate UR-VC on real bimanual cloth flatten-and-fold data.
Since true physical progress is not directly observed in real demonstrations, we do not claim access to ground-truth progress labels.
Instead, our evaluation follows the requirements implied by the method: cross-episode state recurrences should be sufficiently dense, retrieved frames should correspond to the same semantic task state rather than superficial appearance, the corrected labels should express non-monotone progress, and the resulting signal should be usable in a downstream robot-learning pipeline.

\subsection{Experimental Setup}
\label{sec:exp_setup}

\textbf{Task and data.}
We study a long-horizon bimanual cloth flatten-and-fold task.
Each episode starts from a garment placed on a table and requires the robot to flatten the cloth and then fold it into the target configuration.
This task is a natural setting for evaluating time-derived progress proxies: progress is visually meaningful, intermediate states recur across garments and episodes, and local regressions frequently occur due to slips, failed grasps, wrinkles, or partial undoing.
Unless otherwise stated, intrinsic analyses use a primary evaluation set of \num{150} real robot episodes sampled from a flatten-and-fold dataset released by \cite{yu2026chi}.
We additionally report scaling trends by increasing the retrieval pool up to the \(10^4\)-episode scale.

\textbf{UR-VC implementation.}
For each frame, we compute an \(L_2\)-normalized SigLIP-2 visual embedding and retrieve semantically similar frames from other episodes.
Unless otherwise stated, we use a temporal band \(\tau=0.3\) and a cosine-similarity threshold \(\rho=0.90\), which provide a conservative locality constraint and filter weak cross-episode matches.

\textbf{Metrics.}
We evaluate four aspects of the correction.
First, \emph{coverage} measures the fraction of query frames that have at least one valid match from a different episode.
Second, \emph{semantic consistency} checks whether retrieved frames preserve task state across appearance variation.
Third, \emph{non-monotone recovery} measures whether the corrected estimate produces negative horizon advantages in states where progress locally regresses, while preserving the overall task trend.
Finally, we test a downstream use case by using UR-VC-derived advantage labels to train an advantage-conditioned VLA and evaluate it on a dual 7-DoF AgileX robot with absolute joint control.

\begin{figure}[t]
\centering
\includegraphics[width=\linewidth]{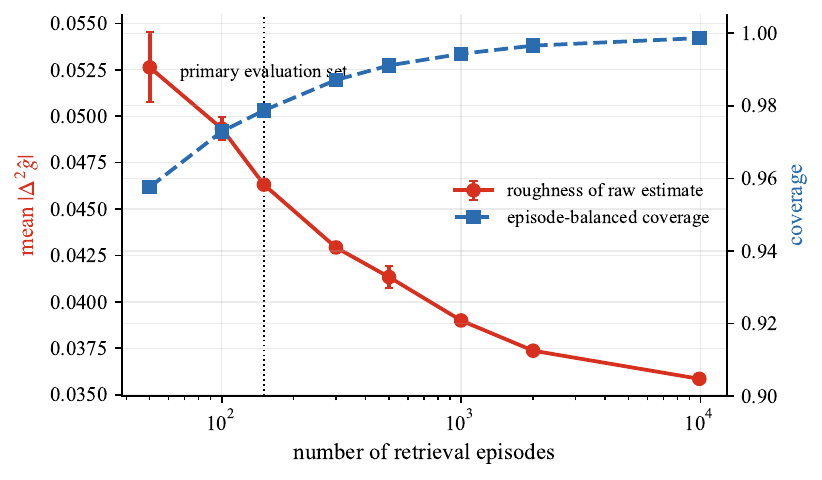}
\caption{\textbf{The correction improves with more retrieval episodes.}
Episode-balanced UR-VC across retrieval-set sizes, reported as mean \(\pm\) std over random subsets.
The roughness of the corrected progress estimate, measured by mean \(|\Delta^2 \hat g_t|\), decreases by roughly one third from \num{50} episodes to the \(10^4\)-episode scale. 
Meanwhile, coverage remains high and rises from \num{96}\% to \num{99.9}\%, showing that larger retrieval sets provide both denser support and more stable correction. The \num{150}-episode primary evaluation set already lies in a usable regime.
}
\label{fig:smooth}
\end{figure}

\subsection{Cross-Episode Coverage and Scaling}
\label{sec:coverage}

UR-VC is useful only if query states have high-quality matches in other episodes.
This condition already holds at the scale of the primary evaluation set.
With \num{150} episodes, \num{98}\% of frames have a nearest neighbor in a different episode with cosine similarity at least \num{0.90}; \num{69.9}\% remain covered under the stricter threshold \num{0.955}.

Coverage further improves with retrieval-set size.
As the retrieval pool increases from the primary set to the \(10^4\)-episode scale corpus, coverage rises to \num{99.9}\% at threshold \num{0.90} and \num{90.4}\% at threshold \num{0.955} (Fig.~\ref{fig:matching}).
This suggests that cross-episode redundancy is not only present in large-scale robot datasets but is already available at moderate collection scales, and it becomes denser as more demonstrations are added.

\subsection{Semantic Consistency of Retrieved States}

High coverage alone is insufficient: visually similar frames must also correspond to similar task states.
Otherwise, averaging their time-derived labels would blur together unrelated configurations rather than correct temporal distortion.
We therefore inspect nearest-neighbor retrievals across different episodes and garments (Fig.~\ref{fig:montage}).
The retrieved frames preserve the folding state across garment colors and appearances, indicating that the embedding primarily captures cloth configuration rather than superficial texture.
These results support the main assumption behind UR-VC: semantically similar task states recur across independently collected episodes, and retrieval can identify them despite appearance variation.

\subsection{Recovering Non-Monotone Progress}

We next examine whether UR-VC expresses progress patterns that normalized time cannot represent.
A normalized-time label is monotone by construction, so its fixed-horizon advantage is non-negative and nearly constant within an episode.
It therefore cannot distinguish an action that improves the cloth state from one that stalls or temporarily undoes progress.

UR-VC preserves the global task trend while allowing local regressions.
In the real episode shown in Fig.~\ref{fig:teaser}, the corrected estimate decreases when a slipping grasp reintroduces wrinkles and the cloth state visibly worsens, whereas the raw time-derived label continues to increase.
Across the primary evaluation set, \num{13.4}\% of frames receive a negative horizon advantage under UR-VC, using (\(r_i<-0.02\) with a horizon of 5\% of the episode length, approximately \(\approx\)1.7\,s).
At the same time, the corrected estimate remains highly correlated with normalized time overall (\num{0.98}), which is expected because the temporal band constrains the correction to remain local.
Thus, UR-VC does not discard the coarse temporal ordering of the demonstration; it selectively corrects local regions where the visual state suggests that progress has regressed.

The correction also becomes more stable as the retrieval pool grows.
From \num{50} episodes to the \(10^4\)-episode scale, the roughness of the corrected estimate, measured by mean \(|\Delta^2 \hat g_t|\), decreases by roughly one third, while coverage increases from \num{96}\% to \num{99.9}
\% (Fig.~\ref{fig:smooth}).
Larger retrieval sets provide both denser support and smoother estimates.

\subsection{Downstream Real-Robot Evaluation}
\label{sec:downstream}

UR-VC produces a corrected progress signal independently of how that signal is consumed.
We evaluate one downstream use case: advantage-conditioned VLA training in the style of \(\pi^{*}_{0.6}\), using a \(\pi_{0.5}\) backbone.

\textbf{Setup.} All methods share the same human-collected training set, model architecture, training schedule, and hyperparameters; one policy is trained per labelling scheme. The training mix combines \num{5700} flatten-and-fold demonstrations with \num{1795} dedicated \emph{recovery} demonstrations that pass through degraded cloth states---precisely the regime where time-derived labels are least reliable. The no-advantage baseline trains without advantage labels. The UR-VC variant computes corrected progress with Eq.~\ref{eq:vc}, marks its top-\num{20}\% advantage frames as positive (Sec.~\ref{sec:application}), and conditions the policy on the resulting positive/negative labels.

\textbf{Evaluation protocol.}
We evaluate on an AgileX bimanual robot performing flatten-then-fold on short-sleeve garments.
The test set contains six table conditions, including a bare-table condition and five tablecloth backgrounds.
For each condition, we evaluate 6 garments with 5 repetitions each, yielding \num{30} trials per condition and \num{180} trials per method.
We report per-condition and the average success rates in Tab.~\ref{tab:real}.

\textbf{Results.}
UR-VC advantage conditioning improves average success from \num{0.728} (131/180) to \num{0.789} (142/180), with higher success in 5 of 6 table conditions.
These results provide application-level evidence that UR-VC-derived advantage labels can improve policy performance when inserted into a standard advantage-conditioned VLA pipeline. Importantly, the improvement is obtained without changing the backbone, training data, optimization schedule, or policy objective, suggesting that the gains are attributable to the supervision signal rather than additional model capacity or training changes.

\begin{table}[t]
\centering
\caption{
\textbf{Downstream real-robot evaluation.}
Conditions correspond to different tablecloth backgrounds, with \num{30} trials per condition. UR-VC improves the average success rate from \num{72.8}\% to \num{78.9}\% and performs better in 5 of 6 conditions.
}
\label{tab:real}
\begin{tabular}{lcc}
\toprule
Condition & Baseline & UR-VC \\
\midrule
Bare table         & 0.90 & \textbf{0.97} \\

Beige cloth        & 0.70 & \textbf{0.73} \\

Blue-gray cloth    & \textbf{0.50} & 0.43 \\

Light-yellow cloth & 0.63 & \textbf{0.90} \\

Light-gray cloth   & 0.77 & \textbf{0.80} \\

Khaki cloth        & 0.87 & \textbf{0.90} \\
\midrule
Average             & 0.728 & \textbf{0.789} \\
\bottomrule
\end{tabular}
\end{table}

\section{Limitations}
\label{sec:limitations}

UR-VC assumes that visually similar observations indicate similar task progress. 
This assumption is shared with progress estimator and value models: when visually similar states occur at different task stages or after different action histories, their latent progress can be ambiguous.
Episode-balanced averaging reduces trajectory-specific timing noise, but it cannot correct systematic retrieval errors.


Our downstream evaluation focuses on one representative use case: using UR-VC-derived supervision for advantage-conditioned VLA training in real bimanual cloth manipulation. Under matched experimental settings, the corrected labels lead to higher average success, suggesting their practical utility as a drop-in supervision signal. Future work may further examine the method across broader task families, training scales, and hyperparameter settings.

\section{Conclusion}
\label{sec:conclusion}

UR-VC corrects time-derived progress proxies by averaging semantic state recurrences across independent robot episodes.
It exploits dataset redundancy to better align scalable proxy labels with latent physical progress.
On real bimanual cloth data, cross-episode matches are abundant and semantic, corrected estimates recover regressions that a monotone time proxy cannot express, and the resulting signal can be used for advantage-conditioned VLA training.
Time provides useful and scalable supervision, but it should not be conflated with physical progress.
As a model-agnostic preprocessing step, UR-VC refines timestamp-derived labels before downstream model training, without requiring an additional value model or any changes to the model architecture or training objective.

\bibliographystyle{plainnat}
\bibliography{references}

\end{document}